\documentclass{article}

\usepackage{arxiv}

\usepackage[utf8]{inputenc} 
\usepackage[T1]{fontenc}    
\usepackage{hyperref}       
\usepackage{url}            
\usepackage{xcolor}
\usepackage{booktabs}       
\usepackage{amsfonts}       
\usepackage{nicefrac}       
\usepackage{microtype}      
\usepackage{lipsum}
\usepackage{graphicx}
\graphicspath{ {./figs/} }
\usepackage{algorithmic}  
\usepackage{algorithm}    
\usepackage{float} 
\usepackage{relsize}
\usepackage{subfigure}
\usepackage{comment}

\usepackage{amsmath}
\usepackage{amssymb}
\usepackage{mathtools}
\usepackage{amsthm}
\usepackage{bbm}
\usepackage[capitalize,noabbrev]{cleveref}

\theoremstyle{plain}
\newtheorem{theorem}{Theorem}[section]
\newtheorem{proposition}[theorem]{Proposition}
\newtheorem{lemma}[theorem]{Lemma}

\newtheorem{example}{Example}
\theoremstyle{definition}
\newtheorem{definition}[theorem]{Definition}

\theoremstyle{remark}

\begin{document}
\title{Do We Truly Need So Many Samples? \\Multi-LLM Repeated Sampling Efficiently Scales Test-Time Compute}

\author{
  \textbf{Jianhao Chen}$^{1~2~\ast}$ \quad 
  \textbf{Zishuo Xun}$^{2~3~\ast}$  \quad
  \textbf{Bocheng Zhou}$^{2~\ast}$ \quad
  \textbf{Han Qi}$^{2~\ast}$ \\
  \textbf{Hangfan Zhang}$^{4}$ \quad
  \textbf{Qiaosheng Zhang}$^{2}$ \quad
  \textbf{Yang Chen}$^{3}$ \quad
  \textbf{Wei Hu}$^1$ \\ 
  \textbf{Yuzhong Qu}$^{1~\dag}$ \quad
  \textbf{Wanli Ouyang}$^2$ \quad 
  \textbf{Shuyue Hu}$^{2~\dag}$ \quad \\
  $^1$ \text{State Key Laboratory for Novel Software Technology, Nanjing University} \\
  $^2$ \text{Shanghai Artificial Intelligence Laboratory}  \\
  $^3$ \text{The University of Auckland} \\
  $^4$ \text{The Pennsylvania State University} \\
  \texttt{jhchen.nju@gmail.com} \quad 
  \texttt{\{whu,yzqu\}@nju.edu.cn} \quad
  \texttt{hbz5148@psu.edu} \\
  \texttt{\{zhoubocheng,qihan,zhangqiaosheng,ouyangwanli,hushuyue\}@pjlab.org.cn} \\
  \texttt{zxun233@aucklanduni.ac.nz} \quad
  \texttt{yang.chen@auckland.ac.nz} \\
}

\renewcommand{\thefootnote}{}

\footnotetext{\dag~Corresponding author.}
\footnotetext{$\ast$~Work done during the author's intership at Shanghai Arificial Intelligence Laboratory.}
\footnotetext{Code and data are available at \url{https://github.com/JianhaoChen-nju/ModelSwitch}.}

\maketitle
\vspace{-20pt}
\begin{abstract}
This paper presents a simple, effective, and cost-efficient strategy to improve LLM performance by scaling test-time compute. 
Our strategy builds upon the repeated-sampling-then-voting framework, with a novel twist:  incorporating multiple models, even weaker ones, to leverage their complementary strengths that potentially arise from diverse training data and paradigms.
By using consistency as a signal, our strategy dynamically switches between models. 
Theoretical analysis highlights the efficiency and performance advantages of our strategy. Extensive experiments on seven datasets demonstrate that our strategy not only outperforms self-consistency and state-of-the-art multi-agent debate approaches, but also significantly reduces inference costs. Additionally, ModelSwitch requires only a few comparable LLMs to achieve optimal performance and can be extended with verification methods, demonstrating the potential of leveraging multiple LLMs in the generation-verification paradigm.

\end{abstract}

\section{Introduction}
\label{Sec:Introduction}
Scaling has been a major driving force of recent rapid advancements in large language models (LLMs). 
While scaling training-time compute~\cite{rae2021scaling} appears to be hitting a plateau, scaling inference-time compute stands out as a promising alternative~\cite{snell2024testtimecompute}.
An emerging direction is to scale inference-time compute based on the \emph{generation-verification} paradigm. 
By querying an LLM with the same question for multiple times,  a number of samples (or candidate answers) are generated, and then these samples are verified to deliver a final answer. 
Studies across various LLMs and benchmarks consistently demonstrate that simply scaling the number of generated samples significantly improves the \emph{coverage} of correct answers~\cite{brown2024large}. Thus, it is perhaps unsurprising that recent attempts have pushed the number of samples to the scale of hundreds or even  thousands~\cite{brown2024large,chen2024llmcalls,liu2025can}, in pursuit of improved answer correctness.

 
However, \emph{do we truly need so many samples}? 
Scaling repeated sampling is undeniably computationally expensive, with the consumption of floating point operations (FLOPs) increasing linearly with the number of samplings~\cite{kaplan2020scalinglaws}. Additionally, in terms of user experience, repeated sampling often leads to significant delay in providing final answers~\cite{HCI}, and no one enjoys waiting too long for a response from AI. Therefore, improving \emph{sample efficiency} is of paramount importance, and there is a pressing need for methods that can deliver correct final answers while minimizing the number of samples required. Recent approaches have primarily focused on the verification side---a great number of outcome or process reward models~\cite{cobbe2021gsm8k,yang2024qwen2,zhang2024generative} and automatic verifiers~\cite{li2022competition,schick2023toolformer} have been proposed, whereas LLM-as-a-judge~\cite{zheng2023judging,lifshitz2025multi} has also been extensively explored.  

Orthogonal to recent efforts, in this paper, we focus on the generation side and explore the potential of leveraging multiple LLMs to improve sample efficiency.  
We argue that employing multiple LLMs for generation can achieve effective complementary capabilities among the models. Trained on different corpora and using distinct paradigms, LLMs exhibit diverse capabilities---even on the same benchmark, two general-purpose LLMs may excel at answering different types of questions~\cite{zhang2025if,wang2024mixture}. 
We test our argument by building upon the simple repeated-sampling-then-voting strategy, following the \emph{Occam's Razor} principle, and present a novel method named \emph{ModelSwitch}. This method introduces two novel twists: (i) incorporating multiple models, even weaker ones, to produce more diverse samples, and (ii) using consistency as a signal to switch models and save compute. The rationale is based on our empirical observation: across various types of LLMs and datasets, their accuracy is positively correlated with the consistency of their generated answers. When a model generates chaotic answers, it serves as a signal to switch models. If the switched model generates consistent answers, there is a higher likelihood of obtaining the correct answer.

\begin{figure}[tbh!]
    \centering
    \includegraphics[width=0.8\columnwidth]{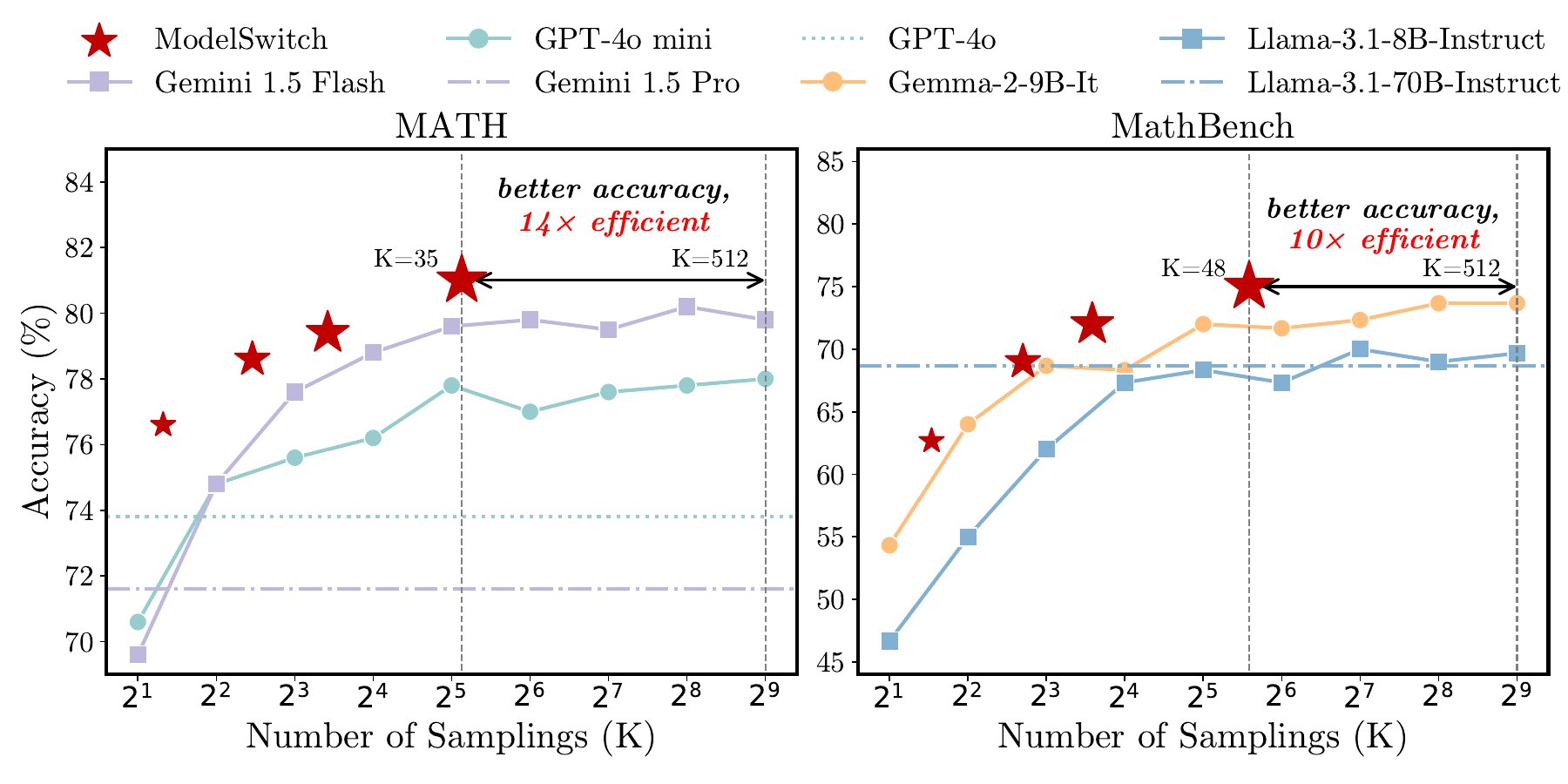}
    \caption{Performance comparison of ModelSwitch and self-consistency~\cite{wang2022self} on Math~\cite{lightman2023let} and MathBench~\cite{liu2024mathbench} dataset. ModelSwitch switches between Gemini 1.5 Flash and GPT-4o mini on MATH, and between Gemma-2-9B-It and Llama-3.1-8B-Instruct on MathBench. The curves illustrate the performance of individual LLMs under self-consistency. For comparison, horizontal lines mark the single-sample performance of larger LLMs, including GPT-4o, Gemini 1.5 Pro, and Llama-3.1-70B-Instruct, as baselines. On MATH, ModelSwitch achieves 81\% accuracy with only 35 samples, outperforming Gemini 1.5 Flash (79.8\% accuracy with 512 samples) while being 14$\times$ more efficient. On MathBench, similar results are observed with open-source models: ModelSwitch achieves 75\% accuracy (48 samples), outperforming Gemma-2-9B-It (73.7\%, 512 samples) with 10$\times$ efficiency. Additionally, combining 9B and 8B models achieves performance (69\%) comparable to a 70B model (68.7\%) with only 7 samples.}
    \label{fig:fig1}
\end{figure}

We conduct extensive evaluations on seven datasets that cover a wide range of knowledge, reasoning, comprehension, and domain-specific challenges. ModelSwitch, when leveraging Gemini 1.5 Flash and GPT-4o mini, significantly surpasses the self-consistency~\cite{wang2022self} performance of each respective LLM. Moreover, it even outperforms more advanced LLMs such as GPT-4o and Gemini 1.5 Pro—achievements that self-consistency cannot match.
At the same time, ModelSwitch demonstrates superior sample efficiency, reducing the average sampling count per query by 34\% on six datasets, showcasing its ability to save computational costs while maintaining high performance. Furthermore, ModelSwitch achieves state-of-the-art performance on four datasets, outperforming five other multi-agent or multi-LLM methods, and secures the second-best result on the remaining two datasets.
Notably, ModelSwitch achieves an impressive 63.2\% accuracy on the MMLU-Pro~\cite{wang2024mmlu} dataset, surpassing the best single LLM by 10.2 points and significantly exceeding other methods, including MAD~\cite{du2023improving} (47.6\%), AgentVerse~\cite{AgentVerse} (44.8\%), ChatEval~\cite{chan2023chateval} (44\%), MAD (multi-LLM version) (44\%), and MOA~\cite{wang2024mixture} (52.6\%).
In additional experiments, we demonstrate that ModelSwitch requires only a small number of LLMs with comparable performance to achieve optimal results and is relatively robust to the order of models. We also show that ModelSwitch can be integrated with verification methods, such as reward models, further enhancing the performance.

We  conduct a formal analysis to better understand the improvements brought by ModelSwitch, compared to self-consistency.
For a given query, we derive a sufficient condition  and a necessary condition that  ModelSwitch (two models) can surpass the performance of  single model. This sufficient condition may still be satisfied  even when neither of the two models can produce a completely consistent answer lists. This reveals the source of the superiority of ModelSwitch compared to a single model. 
Furthermore, we provide a theoretical bound on the efficiency gains achieved by ModelSwitch, offering deeper insights into its advantages.  In particular, if the probability of each model producing a completely consistent answer list is greater than some constant $c>0$ and each model is sampled the same number of times, the expected number of samplings can be decreased by a factor $\frac{1}{n\times c} $, where $n$ is the number of models. 

In summary, our key contributions are outlined as follows:
\begin{enumerate}

\item An empirical analysis that reveals a universal correlation between consistency and accuracy of generated answers across various popular LLMs and datasets.

\item A simple, effective, and cost-efficient generation-verification method that effectively leverages the complementary strength of multiple LLMs.

\item Extensive experiments demonstrating that ModelSwitch outperforms single-LLM sampling-then-voting in efficacy and efficiency, and more effectively leverages multi-LLM synergies compared to debate-based approaches.

\item A theoretical analysis that provides a deeper understanding of why multiple-LLM generation is superior to single-LLM generation. 
\end{enumerate}

\section{A Universal Correlation between Consistency and Accuracy}
\label{sec:correlation}
In this section, we conduct an empirical analysis of the relationship between consistency (in terms of the entropy of the generated answers) and accuracy of multiple popular LLMs. Self-consistency~\cite{wang2022self} has observed that the consistency (here measured as the percentage of decodes agreeing with the final aggregated answer) is highly correlated with accuracy on the GSM8K dataset. However, this observation was limited to a single dataset and a single model. To verify the generality of this finding, we extend our analysis to multiple datasets and several mainstream LLMs. Specifically, we use entropy to quantify the consistency among different generated answers. Compared to the  percentage of decodes agreeing with the final aggregated answer, entropy provides a more comprehensive representation as it provides a more accurate characterization of the generated answer distribution.
\begin{figure*}[!h]
    \centering
    \includegraphics[width=\textwidth]{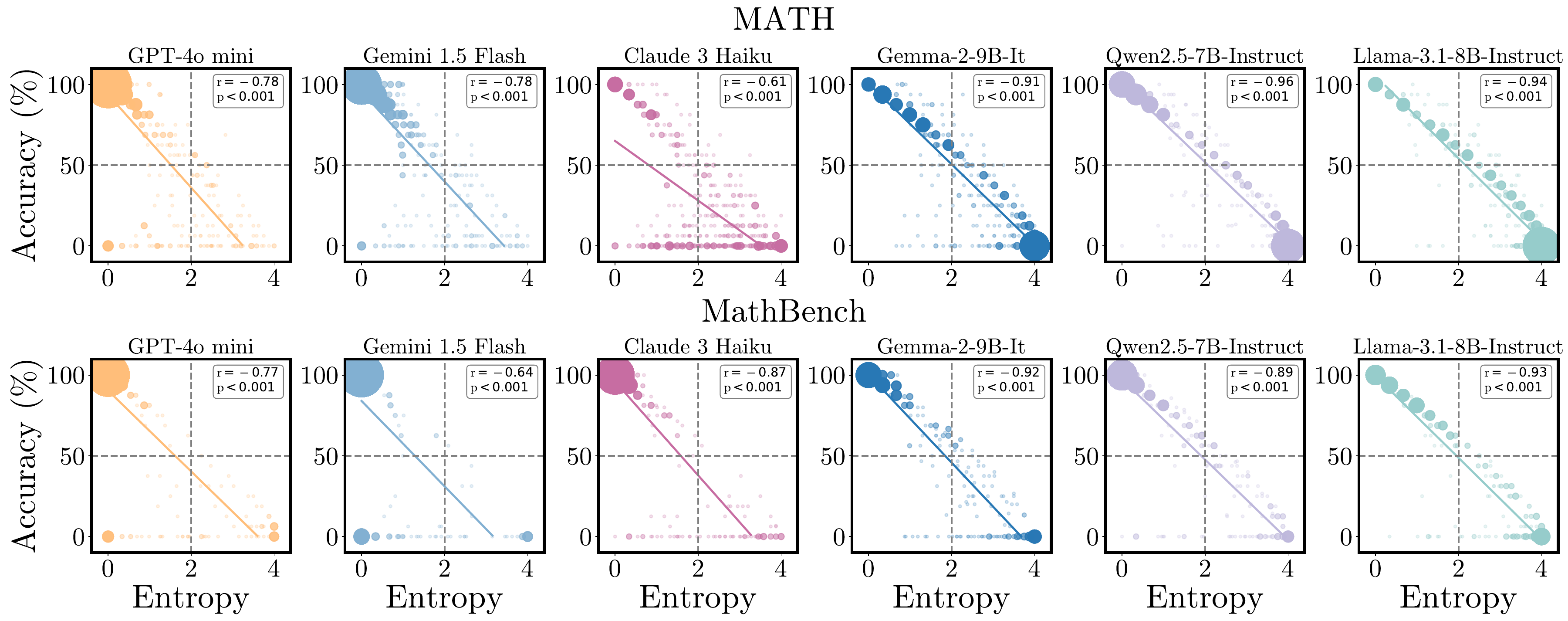}
    \vspace{-10pt}
    \caption{Correlation between consistency (entropy) and accuracy of answers from six LLMs on MATH and MathBench. We use the transparency and size of the scatter points to indicate the number of queries corresponding to each point (the larger and more opaque the point, the greater the quantity). We report the correlation coefficient $r$ and the significance indicator $p$. In each subplot, entropy and accuracy exhibit a moderate ($0.5 < |r| \leq 0.8$) or high ($|r| > 0.8$) correlation, and this correlation is statistically significant ($p < 0.001$). }
    \label{fig:MB_Correlation}
\end{figure*}

As shown in Figure~\ref{fig:MB_Correlation}, we fix each LLM to sample 16 times per query, and calculate the entropy and  the accuracy of the 16 answers.  Consistency (entropy) and accuracy exhibit a strong correlation (r), and this trend is universal across all six LLMs, irrespective of their different sizes and performances.  There are three extreme cases that can be discussed separately: those distributed in the top-left (high consistency, high accuracy), bottom-left (high consistency, low accuracy), and bottom-right (low consistency, low accuracy). Distribution in the bottom-right indicates that the model generated 16 entirely different answers, suggesting a lack of confidence in the current query. In such cases, even if a correct answer exists, it is difficult to be selected without oracle verifiers. Distribution in the top-left and bottom-left means the model generated only one type of answer, indicating high confidence. However, the difference is that the former is confidently correct, while the latter is confidently wrong. The points in the top-left greatly outnumber those in the bottom-left, which means that when an LLM generates highly consistent answers for a query, there is a strong likelihood that the answer is correct. \emph{In a word, consistent implies correct, while chaotic implies wrong}.

\section{ModelSwitch: Harnessing Consistency in Multi-LLM Generation}
\begin{figure*}[!h]
    \centering
    \includegraphics[width=\textwidth]{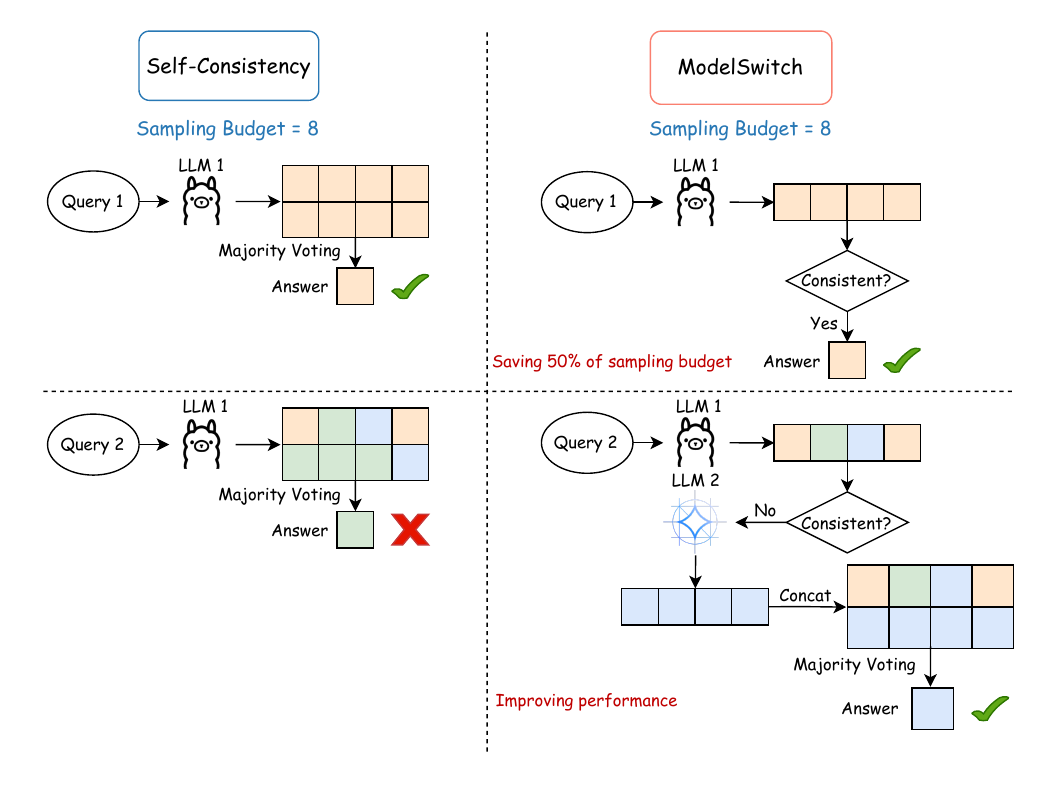}
    \vspace{-20pt}
    \caption{An overview of how ModelSwitch approach works between two LLMs.  Given a sample budget $K$, it first queries the first LLM with $\frac{K}{2}$ samples. If self-consistency is achieved, the answer is accepted, saving 50\% of the budget. Otherwise, it queries the second LLM with left $\frac{K}{2}$ samples and aggregates answers from both models, potentially improving performance if the second LLM excels at the task.}
    \label{fig:frame}
\end{figure*}

\emph{ModelSwitch} is a simple yet effective framework for multi-LLM repeated sampling  that simultaneously achieves high performance and efficiency. Based on the analysis in Section~\ref{sec:correlation}, our core idea is to leverage the strong correlation between consistency and accuracy during repeated sampling. When consistency is high, we can reduce the number of samplings and confidently rely on the current answer. Conversely, when consistency is low, it indicates that the model "does not know" the answer. In such cases, rather than persisting with the same model, we can switch to another model, embracing the possibility that "the next model you encounter might know what the previous one did not." This approach enables complementary strengths across models, optimizing both efficiency and accuracy. Combining  these insights, we present our ModelSwitch design, detailed in Algorithm~\ref{alg:model_switch}.
\begin{algorithm}
   \caption{ModelSwitch Algorithm}
   \label{alg:model_switch}
\begin{algorithmic}
   \STATE {\bfseries Input:} Model set $\{M_1, M_2, \ldots, M_n\}$, sampling budget = $K$
   \STATE {\bfseries Output:} $Answer$
   \STATE Initialize $All\_M.answers = \{\}$
   \FOR{$i = 1$ {\bfseries to} $n$}
      \STATE Initialize $M_i.answers = \{\}$
      \FOR{$j = 1$ {\bfseries to} $\frac{K}{n}$}
         \STATE $sample = M_i.\text{sampling}()$
         \STATE $answer$ = Extract\_Answer($sample$)
         \STATE Add $answer$ to $M_i.answers$
      \ENDFOR
      \IF{$ i \leq n-1$ and size(set$(M_i.answers))==1$}
         \STATE $Answer=M_i.answers(0) $ and Exit
      \ENDIF 
      \STATE Add $M_i.answers$ to $All\_M.answers$
   \ENDFOR
   \STATE $Answer = \text{voting\_algorithm}(all\_results)$
\end{algorithmic}
\end{algorithm}

ModelSwitch replaces a single LLM with multiple LLMs during the sampling stage. First, the selection of LLMs plays a crucial role in the effectiveness of this approach. We prioritize diverse models over different versions of the same one. This is because similar versions often share overlapping knowledge, which limits diversity and reduces the potential for complementary insights. The next step in this approach is determining how to allocate the sampling budget across the models. In our method, we simply set the sampling budget for each LLM to be equal. This means that, given a fixed total sampling budget, instead of a single model sampling $K$ times, we distribute the budget evenly across $n$ models, allowing each to sample $\frac{K}{n}$ times without increasing the total sampling budget. Empirically, when performing ModelSwitch, it is best to arrange the models from strongest to weakest overall. If a model earlier in the order can answer the current query, the remaining LLM calls are saved, which improves computational efficiency.

\paragraph{Weighted Voting Algorithm}
To account for performance differences among models, we use a weighted voting algorithm with internal ($\text{W}_\alpha$) and external ($\text{W}_\beta$) weights to aggregate answers $A$ from all models $M$.
\begin{equation}
MS(q) = \operatorname*{argmax}_{a \in A} \sum_{i=1}^{M} \text{W}_\alpha \times\text{W}_\beta \times \text{count}(a, A_i)
\end{equation}
Internal Weights: Measure a model's confidence based on answer consistency. Calculated via entropy and normalized to $[\text{bias}, 1]$, where $\text{bias} = \frac{1}{\text{len(samples)}}$. Higher consistency yields a higher weight.  Example: For answer sets $\text{M}_1 = \{\text{A} \times 3, \text{B} \times 2\}$, $\text{M}_2 = \{\text{B} \times 3, \text{A} \times 2\}$, $\text{M}_3 = \{\text{C} \times 5\}$, $\text{M}_3$'s uniformity gives $\text{C}$ a higher score.
{\small 
\begin{align}
\label{formu:weight}
\text{W}_\alpha = 
\text{bias} + (1 - \text{bias}) \left(1 - \frac{\text{Entropy}}{\text{norm}}\right), & \text{ if } \text{norm} > 0 \text{ else } 1  
\end{align}
}
External Weights: Assess prior performance, ensuring models with significantly different performances have corresponding importance. Specific $\text{W}_\beta$ settings can be found in~\cref{App:Setup}.

\paragraph{Special Case}
Two-LLM switch is a very special case, as shown in Figure~\ref{fig:frame}, not only because it is easy to set up but also because it has good properties under specific conditions. When the sampling counts for the two models are equal (denoted as $\frac{K}{2}$), the effect of ModelSwitch under the sampling budget $K$ is exactly equivalent to each model sampling $\frac{K}{2}$ completely and then mixing. This means that we achieve completely lossless performance while greatly enhancing efficiency. This property is straightforward to prove: consider two models, $M_1$ and $M_2$, each generating $\frac{K}{2}$ samples. Let $A_1$ be the most frequent answer from $M_1$, with frequency $f(A_1)$. The performance remains lossless because:
\begin{itemize}
    \item If $f(A_1) < \frac{K}{2}$, the samples of both models are considered.
    \item If $f(A_1) = \frac{K}{2}$, $A_1$ is selected, and $M_2$'s samples are pruned without affecting the final answer. Assuming a tie-breaking rule that favors
    $M_1$'s answer.
\end{itemize}

\section{Experiments: On the Efficacy, Efficiency, Scalability, and Robustness of ModelSwitch}
Our experiments aim to answer the following four questions: (i) Does ModelSwitch outperform single-LLM repeated-sampling in terms of efficacy and efficiency? 
(ii) Can ModelSwitch surpass other multi-agent (LLM) debate methods? 
(iii) Does scaling the number of LLMs involved in ModelSwitch lead to continued performance improvements? 
(iv)Can ModelSwitch be extended with stronger verification methods?

\subsection{Experimental Setup}
\paragraph{Benchmarks}
We mainly evaluate our method on the following seven datasets that cover a wide range of knowledge, reasoning, comprehension, and domain-specific challenges, providing a robust evaluation for our method: GSM8K~\cite{cobbe2021gsm8k}, MATH~\cite{lightman2023let}, MathBench~\cite{liu2024mathbench}, MGSM~\cite{shi2022language}, DATE~\cite{wei2022chain}, MMLU-Pro~\cite{wang2024mmlu} and AIME24. Details of the seven datasets are provided in~\cref{App:Setup}.

\paragraph{Models}
We primarily consider three lightweight, closed-source LLMs: GPT-4o mini~\cite{hurst2024gpt}, Gemini 1.5 Flash~\cite{team2024gemini}, and Claude 3 Haiku~\cite{claude}. In Section~\ref{sec:vsSC}, we compare ModelSwitch to single-LLM repeated sampling methods, using GPT-4o mini and Gemini 1.5 Flash, while also including more advanced LLMs, GPT-4o and Gemini 1.5 Pro, as additional baselines. Subsequently, in Section~\ref{sec:vs.debate}, we use all three lightweight LLMs to evaluate ModelSwitch against other multi-agent debate methods. In Section~\ref{sec:llm_number}, we investigate the impact of the number of LLMs by combining the three aforementioned lightweight closed-source LLMs with three open-source LLMs—Llama-3.1-8B-Instruct~\cite{dubey2024llama}, Gemma-2-9B-It~\cite{team2024gemma}, and Qwen2.5-7B-Instruct~\cite{yang2024qwen2}—to assess how performance scales with model diversity. Finally, in Section~\ref{sec:RM}, we explore whether ModelSwitch can be enhanced by integrating stronger validation methods, using Qwen2.5-MATH-RM-72B. \cref{AIME} explores efficacy of ModelSwitch using reasoning LLMs 
Llama-3.1-Nemotron-Nano-8B-v1~\cite{sun2025reward} and DeepSeek-R1-Distill-Qwen-7B~\cite{guo2025deepseek}.






\subsection{ModelSwitch Outperforms Single-LLM Sampling-then-Voting in Efficacy, Efficiency and Scalability}
\label{sec:vsSC}
\begin{figure}[t!]
    \centering
    \includegraphics[width=0.8\textwidth]{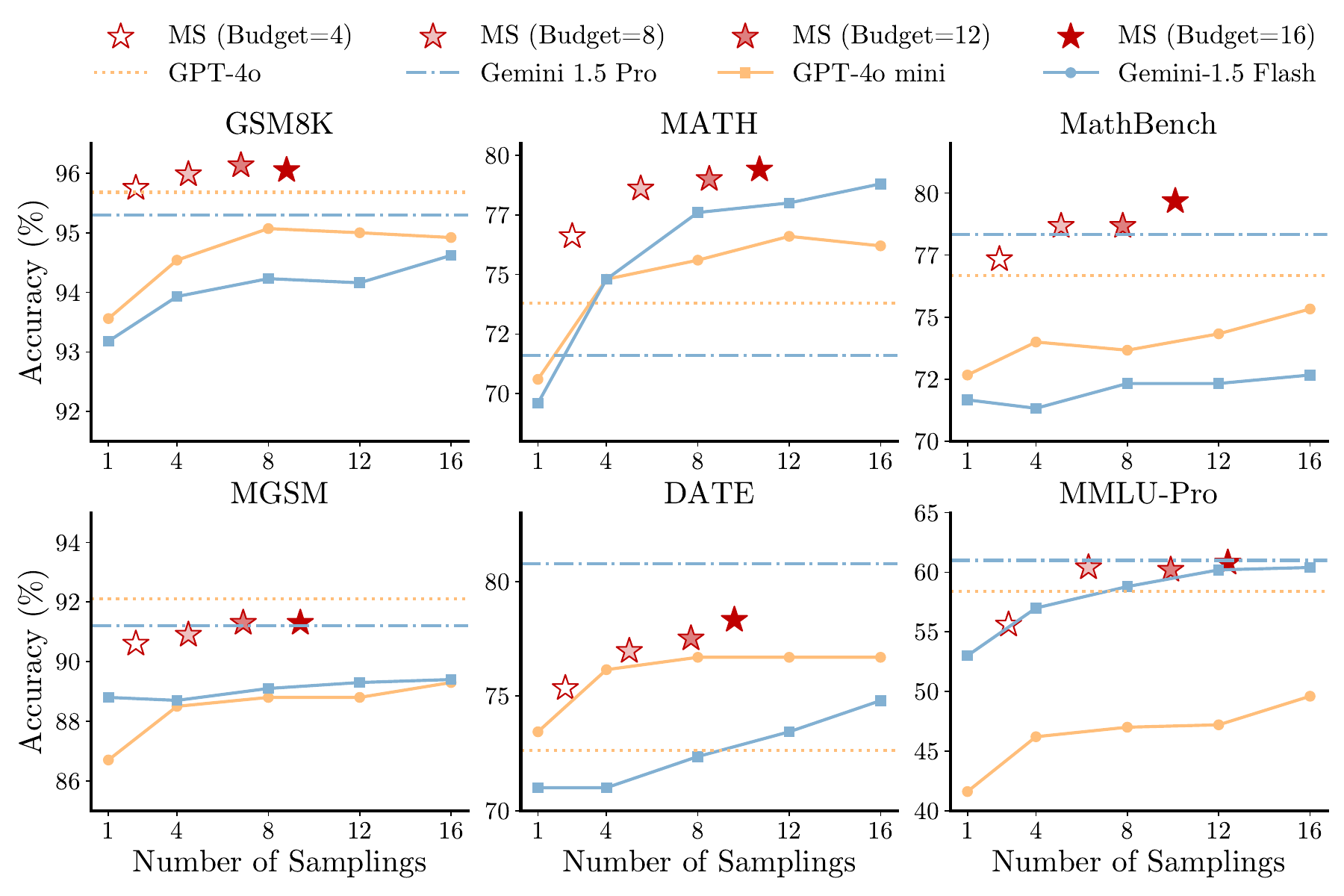}
    \caption{Performance comparison of self-consistency for each LLM (GPT-4o mini and Gemini 1.5 Flash) and ModelSwitch using both. We  use horizontal lines to mark the single-sample results of more advanced LLMs GPT-4o and Gemini 1.5 Pro.   We use the shade and size of red stars to differentiate the sampling budgets of ModelSwitch. The horizontal coordinate of the red star reflects the actual sampling counts of ModelSwitch. }
    \label{fig:main_performance}
\end{figure}
In this section, we compare ModelSwitch with self-consistency, which is the classic approach for single-LLM sampling-then-voting.  To ensure fair comparison, we evaluate self-consistency for each LLM (GPT-4o mini and Gemini 1.5 Flash) and ModelSwitch using both, under the same sampling budget. All our queries are asked in COT~\cite{wei2022chain} format by default. The results are shown in Figure~\ref{fig:main_performance}.

\paragraph{Efficacy}
Across all the benchmarks considered, ModelSwitch with two LLMs outperforms single-LLM self-consistency, even when using the best-performing single LLM.
For instance, as the sampling budget increases from 1 to 16, ModelSwitch delivers a 7-point performance boost on MathBench (increasing from 72.7\% to 79.7\%), significantly outperforming the self-consistency gains of Gemini 1.5 Flash at 2.6-point (72.7\% to 75.3\%) and GPT-4o mini at 1-point (71.7\% to 72.7\%). On MMLU-Pro, even when switching between models with a performance gap exceeding 10\%, we can still achieve an improvement over self-consistency of the best-performing single LLM.

\paragraph{Efficiency} 
While ModelSwitch outperforms self-consistency, it also requires fewer samples. 
Taking the sampling budget of 16 as an example, ModelSwitch reduces the average actual sampling count per query to 9.2, 10.7, 10.5, 9.4, 10.6, and 13.4 across the six datasets. (Note that since we calculate the average sampling counts per query across the entire dataset, the actual sampling counts for ModelSwitch may not necessarily be integers.)
\emph{Overall, ModelSwitch saves 34\% samples on average, and saves up to 43\% on GSM8K and 41\% on MGSM, respectively.}

\paragraph{Scalability} ModelSwitch leveraging GPT-4o mini and Gemini 1.5 Flash, exhibits superior scalability compared to self-consistency for each LLM, outperforming more advanced LLMs GPT-4o and Gemini 1.5 Pro. For example, with an average of 2.2 samples per query on GSM8K, ModelSwitch surpasses both GPT-4o and Gemini 1.5 Pro. \emph{Similarly, on MathBench, ModelSwitch outperforms both models with an average of 5.1 samples per query—achievements that self-consistency fails to match.}

\subsection{ModelSwitch Better Leverages  Multi-LLM or Multi-Agent Synergies than Debate Methods}
\label{sec:vs.debate}
In this section, we evaluate our approach against other multi-agent debate methods that have demonstrated competitive performance: MAD~\cite{du2023improving}, selected as a canonical representative of the multi-agent debate paradigm, establishes a strong baseline for debate-style collaboration through its innovative use of inter-agent critique to enhance factuality and reasoning capabilities; ChatEval~\cite{chan2023chateval}, chosen for its extension of the MAD framework, showcases how multi-agent collaboration can effectively mimic human evaluation processes, particularly in complex assessment tasks; AgentVerse~\cite{AgentVerse}, which emphasizes dynamic group composition and the emergence of social behaviors among agents, enabling more adaptive and synergistic collaboration in diverse scenarios; and MOA~\cite{wang2024mixture}, which includes a novel hierarchical architecture that aggregates samples from multiple LLM agents engaging in critical thinking through mutual evaluation, and which we perceive as a form of implicit debate.
\begin{figure*}[t!]
    \centering
    \includegraphics[width=0.8\textwidth]{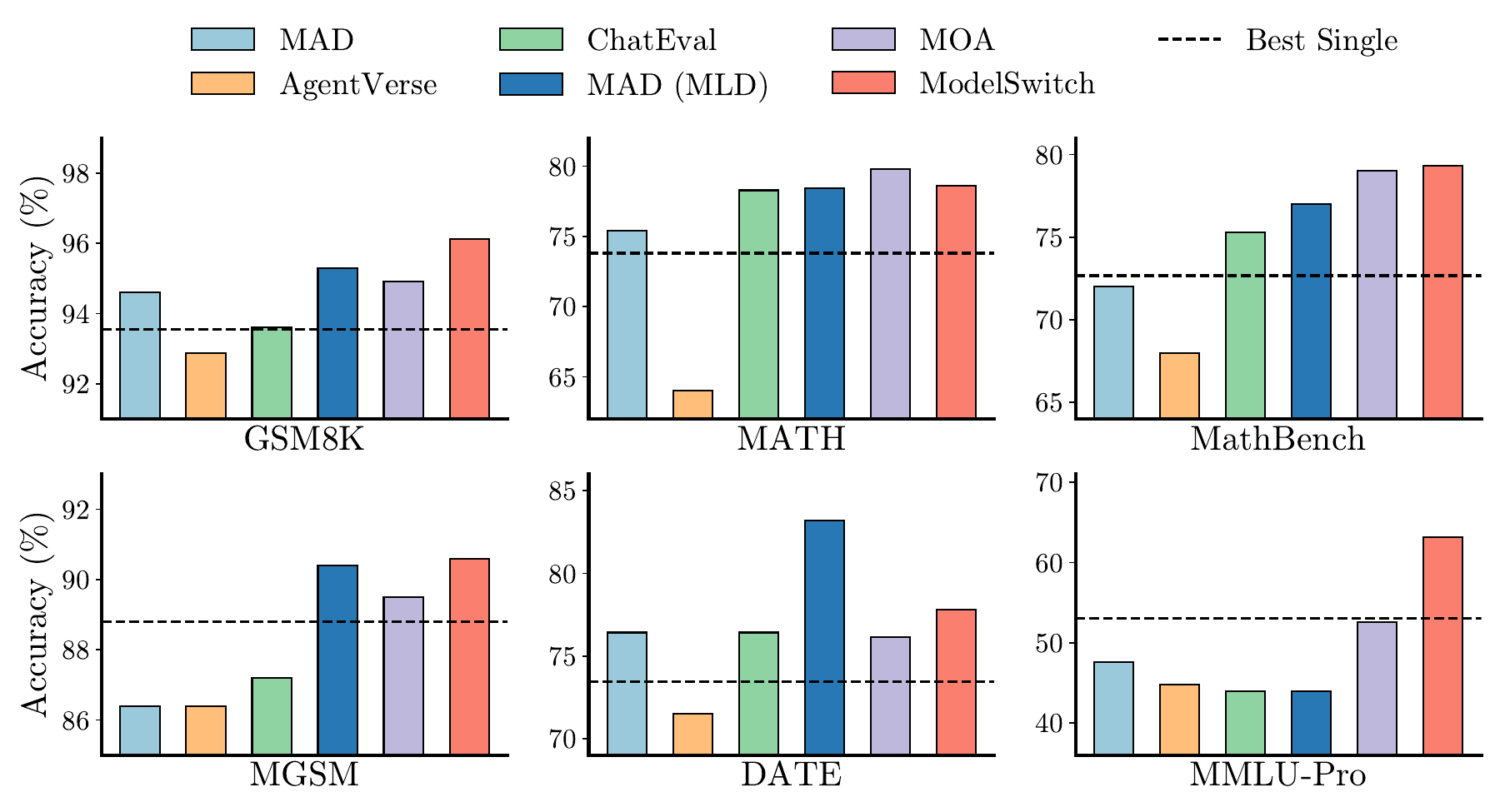}
    \caption{Performance comparison of different multi-agent debate systems under the same budget of 15 samples relative to the accuracy of the best single LLM with one sample. ModelSwitch achieves the best results on four datasets and the second-best result on MATH and DATE.}
    \label{fig:vsMA}
\end{figure*}
For MAD and AgentVerse, we use a single LLM, GPT-4o mini. We additionally construct a new baseline, referred to as MAD (MLD), by extending MAD into a multi-LLM debate version. In ChatEval and MAD (MLD), we conduct debates using GPT-4o mini and Gemini 1.5 Flash. In MOA and ModelSwitch, we utilize GPT-4o mini, Gemini 1.5 Flash, and Claude 3 Haiku.
Due to the varying architectures of different systems, a relatively fair comparison is to unify the sampling budget across all systems when evaluating their performance. 
We set the total number of LLM sampling budget for all systems to 15. For specific settings of different systems, see~\cref{App:Setup}.
\paragraph{Efficacy}
As shown in Figure~\ref{fig:vsMA}, our method achieves state-of-the-art performance on four datasets, with a significant lead on MMLU-Pro. For example, MAD~\cite{du2023improving} (47.6\%), AgentVerse~\cite{AgentVerse} (44.8\%), ChatEval~\cite{chan2023chateval} (44\%), MAD (multi-LLM version) (44\%), and MOA~\cite{wang2024mixture} (52.6\%) all perform worse than the best single LLM (53\%) sampling once on MMLU-Pro. This performance gap highlights a key limitation of these systems: while they rely on interactions between multiple agents (LLMs), it is difficult to ensure that such interactions consistently guide the answers in the correct direction. Challenges like error propagation often arise during these interactions~\cite{zhang2025if,wang2024rethinking}, which are especially hard to mitigate on a complex dataset like MMLU-Pro. In contrast, ModelSwitch achieves a remarkable 63.2\% accuracy, representing an impressive 10.2-point increase over the best single LLM. This demonstrates its ability to effectively address these issues and deliver superior performance. This also highlights the potential of exploring how to effectively promote collaboration among different LLMs as a promising research direction.

\subsection{ModelSwitch Needs Only Few, Comparable LLMs and Remains Robust to LLM Order}
\label{sec:llm_number}
\begin{figure}[t!]
    \centering
    \includegraphics[width=\textwidth]{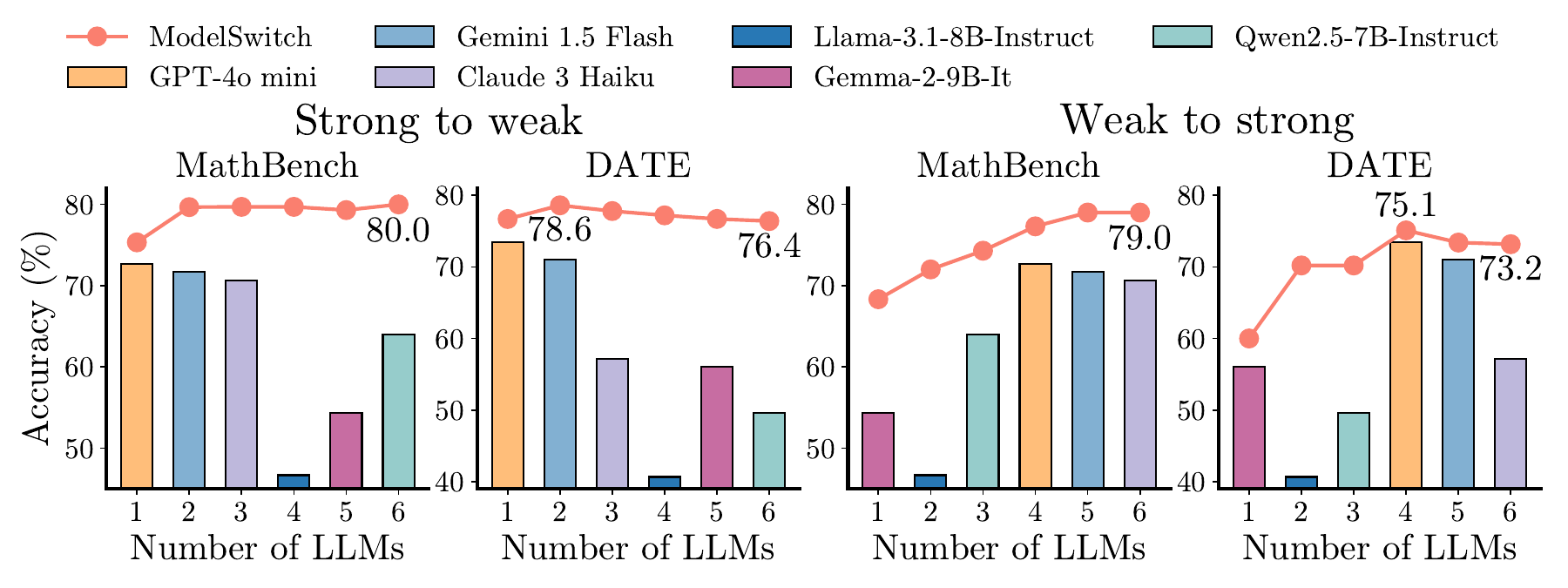}
    \caption{Performance of scaling the number of LLMs under two settings: strong-to-weak and weak-to-strong. The bars show the performance of each individual model with one sample, whereas the curves show the performance of ModelSwitch when involving the models represented by the current and all previous bars. All data in the figure are results obtained under the same sampling budget of 16 samples. In each subplot, we have marked the optimal performance and the performance where all models are included. Few, competing LLMs are sufficient to achieve the optimal performance of ModelSwitch. Moreover, ModelSwitch is relatively robust to order when containing same models.}
    \label{figs：scaling_llm_n}
\end{figure}
In this section, we investigate the impact of the number and order of LLMs on the performance of ModelSwitch. As shown in Figure~\ref{figs：scaling_llm_n}, we experimented with scaling the number of LLMs involved from 1 to 6. To study the effect of order on performance when increasing the number of models, we used two settings: strong-to-weak and weak-to-strong. During this scaling process, we consistently maintained the sampling budget at 16, aiming to demonstrate the potential of expanding along the dimension of model quantity under the same sampling budget. We selected three closed-source LLMs with relatively close performance and three open-source LLMs also with relatively close performance. There is a significant performance gap between the closed-source and open-source LLMs. It would be interesting to observe whether there is a sharp decline in performance when strong models are combined with weaker ones. 

\paragraph{Effects of LLM Number}
Under the strong-to-weak setting, the most significant improvement in performance is observed when scaling from 1 model to 2 models. However, from 2 to 6 models, the performance either plateaued ($79.7\%->80.0\%$ on MathBench) or declined ($78.6\%->76.4\%$ on DATE). This indicates that simply increasing the number of models does not continuously enhance performance. Selecting few, comparable LLMs for ModelSwitch yields the best results.

\paragraph{Effects of LLM Order}
Comparing the strong-to-weak setting and the weak-to-strong setting, the strong-to-weak setting achieves better overall performance unsurprisingly. Notably, on two datasets, the performance of the weak-to-strong arrangement does not decrease significantly ($80.0\% -> 79.0\%$, $76.4\% -> 73.2\%$). The results from the weak-to-strong arrangement also suggests the relative robustness of our method to the order of models.


\subsection{ModelSwitch Can be Combined with Stronger Verification For Performance Boost}
\label{sec:RM}

\begin{figure}[t!]
    \centering
    \includegraphics[width=0.8\textwidth]{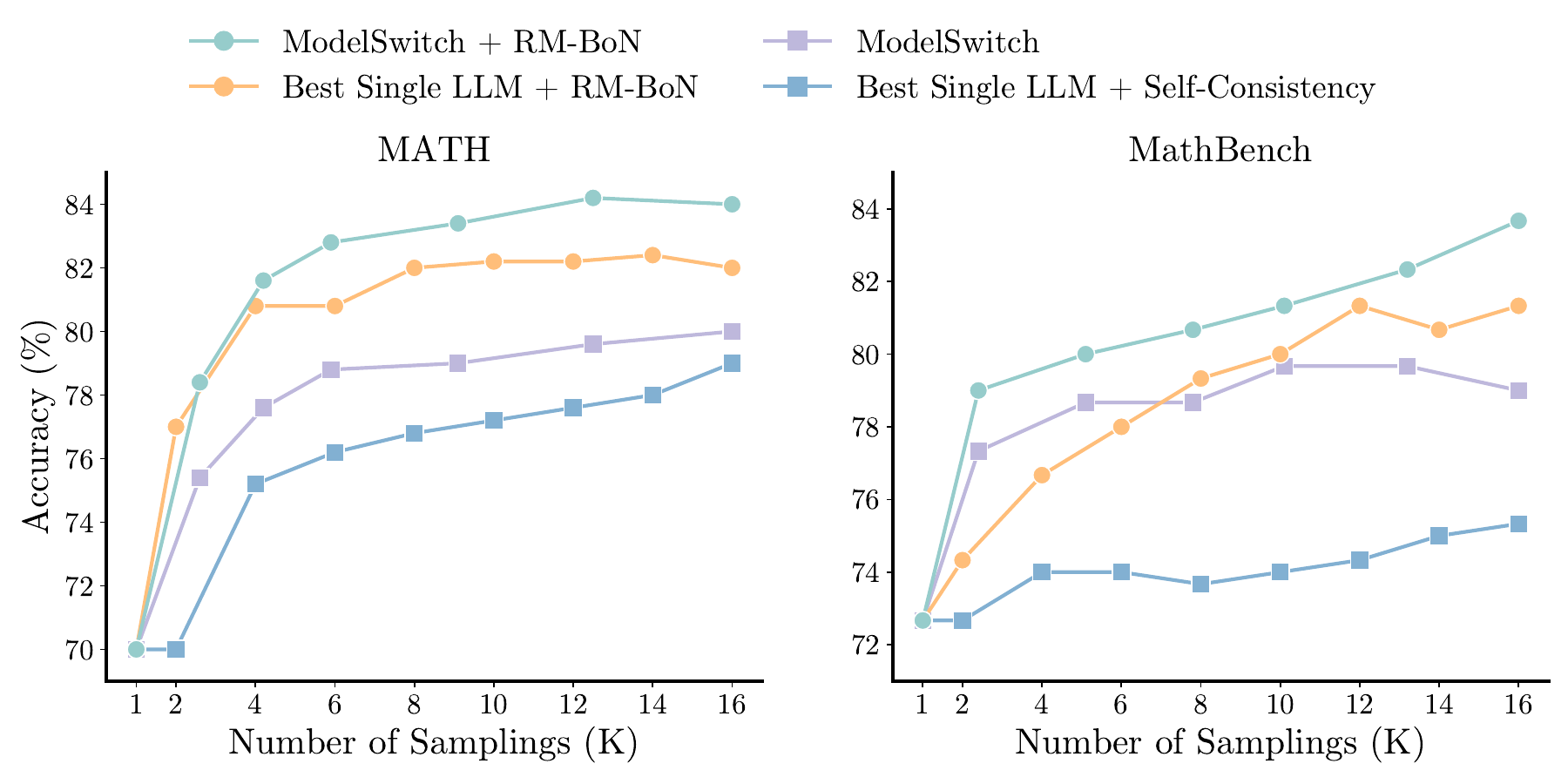}
    \caption{Performance comparison of ModelSwitch and single LLM  combined with Qwen2.5-MATH-RM-72B as the reward model using the Best of N strategy (denoted as RM-BoN). We use "Best Single LLM" to refer to the best-performing model on each dataset: specifically, Gemini 1.5 Flash on MATH and GPT-4o mini on MathBench. ModelSwitch, integrating GPT-4o mini and Gemini 1.5 Flash, achieves superior accuracy on both MATH and MathBench datasets, outperforming the best single LLM under the same RM-BoN strategy.}
    \label{fig:RM}
\end{figure}

In this section, we combine ModelSwitch with Qwen2.5-MATH-RM-72B~\cite{yang2024qwen2}, a highly capable mathematical reward model, to score different samples from GPT-4o mini and Gemini 1.5 Flash. We utilize the Best of N (BoN) strategy to select the answer with the highest score instead of voting . Specifically, our approach is to replace the voting\_algorithm (all\_results) in Algorithm~\ref{alg:model_switch} with BoN (all\_results). We compare the performance curves of the best single LLM and ModelSwitch under RM-BoN strategy. The results are shown in Figure~\ref{fig:RM}.

ModelSwitch, when combined with stronger verification method (RM-BoN), achieves a notable improvement in performance (with 16 samplings, accuracy increases from 80\% to 84\% on MATH and from 79\% to 84\% on MathBench). At the same time, it consistently outperforms the best single LLM repeated sampling + RM-BoN (with 16 samplings, ModelSwitch + RM-BoN achieves 84\% accuracy on MATH vs. 82\% for the best single LLM + RM-BoN, and 84\% on MathBench vs. 81.33\%). Interestingly, on MathBench, vanilla ModelSwitch even surpasses the best single LLM + RM-BoN  given a small number of samplings. Specifically, with a sampling count of 5, ModelSwitch achieves a 78.7\% accuracy, outperforming the best single LLM + RM-BoN with 6 samplings (78\%), demonstrating the efficiency advantages of our method.

Overall, the ModelSwitch approach primarily emphasizes strategies in the generation (sampling) phase, thus it can be further enhanced by stronger selection (verification) methods to improve overall performance. Multi-LLM generation still holds significant potential. 

\section{ Understanding Why ModelSwitch Outperforms Self-Consistency}
\label{section:Theo}
In this section, we analyze when ModelSwitch can improve performance for a given fixed query. For the analysis on the entire dataset, please refer to the \cref{section:Theo}.  To simplify the problem, we only analyze the case of two models $M_1$ and $M_2$. We assume that the first option is the correct answer. Let $p_1=(x_1,...,y_1,...)$ and $p_2=(x_2,...,y_2,...)$  denote the output distributions of $M_1$ and $M_2$ for some fixed query, where  $x_1$ and $ y_1$ are defined as follows (the definition of $x_2$ and $y_2$ are similar): If $x_1$ is the largest one in the probability vector $p_1$, then $y_1$ is the second largest; otherwise, $y_1$ is the largest. 

Assume that the sampling counts for the two models are equal. We focus on the expected performance,  we can directly compare two probability vectors to find out the performance difference between the two mixed models and a single model.
The probability of correct answer of ModelSwitch can be denoted as $P = x_1+x_2$. If $P$ is the largest one in the probability vector $p_1+p_2$, the ModelSwitch  can obtain the correct answer (in the sense of expectation). For situations where both $M_1$ and $M_2$ provide the correct final answer or both give the same incorrect final answer, ModelSwitch does not change the final answer so we do not consider these two situations. Let $Q=y_1 - y_2$ and the number of options is denoted as  $m$.
We have the following proposition.

\begin{proposition}
\label{prop}
     The necessary condition for ModelSwitch to obtain the correct answer is $P>\frac{2}{m}$. 
     The sufficient condition is
     \[ P+Q+x_1>1 \text{ and } P-Q+x_2>1. \]
\end{proposition}
\begin{proof}
    (1) \emph{Sufficiency}: From the definition of $x_1$ and $y_1$ ($x_2$ and $y_2$), $y_1$ and $y_2 $ is the largest or second largest in $p_1$ and $p_2$. To ensure $P$ is the largest one  in the probability vector $p_1 + p_2$, $P$ only needs to be greater than the sum corresponding to $y_1$ and $y_2$. Thus we have, $ P > y_1 + 1-x_2 -y_2 $ and $P > y_2 + 1-x_1-y_1$.

    (2) \emph{Necessity}: If $P \leq \frac{2}{m}$, the average value of the remaining part of $p_1+p_2$ after removing $x_1 + x_2$ is greater than $ \frac{2-2/m}{m-1} = \frac{2}{m}$.  There exists some value  in $p_1+p_2$ that is greater than $\frac{2}{m}$. Hence, ModelSwitch cannot get the correct answer when $P \leq \frac{2}{m}$. 
 
\end{proof}

The sufficient condition in \cref{prop} can be intuitively explained as follows: $ P $  should be as large as possible, which is the direct factor that enables ModelSwitch to yield the correct answer. On the other hand, $ |Q| $ should not be overly large, as this would result in at least one of the inequalities in the sufficient condition failing to hold.  This is because if $ |Q| $ is too large, it implies that the probability of an incorrect answer from a particular model ($M_1$ or $M_2$) has an overwhelming advantage, making it difficult to enhance performance through mixing.

Note that, this proposition does not impose constraints on whether $M_1$ and $M_2$ can obtain the correct  final aggregated answer individually, i.e., it does not require $x_1$ or $x_2$ to be the largest probabilities.  Even both models cannot obtain the correct  final aggregated answer, the above proposition indicates that ModelSwitch  still have chance to obtain the correct  final aggregated answer. This is why repeated sampling with multiple LLMs yield better results than with a single LLM. 

For example, if $p_1 = (0.4,0.6,0)$ and $p_2 = (0.4,0.0.6)$, both $M_1$ and $M_2$ cannot obtain the correct answer, while the mixed model can. This example illustrates that ModelSwitch can reduce the system's confidence in each incorrect answer. Thus,  \textbf{by ModelSwitch, different errors in various models can counteract each other}. For another example, 
if $p_1= (0.7,0.1,0.2,0)$ and $p_2=(0.15,0.05,0.05,0.75)$, $M_1$ can obtain the correct answer but $M_2$ can't. Then the mixed model can output the correct answer. It can be verified that both examples satisfy Proposition \ref{prop}.

\paragraph{Computational Efficiency}
Consider a set of \( n \) models, denoted as \( \{M_1, M_2, \ldots, M_n\} \). For a given query and a total sampling budget \( K \), each model is sampled sequentially up to \( \frac{K}{n} \) times, such that the theoretical maximum number of samples is \( K \). Each model \( M_i \) has a probability \( P_i \) of producing a completely consistent answer list,  where \( P_n  =1\) indicates that if the first \(n-1\) models do not produce a consistent answer, the $n$-th model must be sampled.. If a model produces such a consistent answer list, subsequent models are pruned, and no further sampling is performed.

In this scenario, the \textbf{expected actual number of samplings} (\( \mathbb{E}[N_{\text{call}}] \)) can be bounded as follows:
\begin{equation}
\small
\mathbb{E}[N_{\text{call}}] = \frac{K}{n} \sum_{i=1}^{n} \left( \prod_{j=1}^{i-1} (1 - P_j) \right) P_i
\end{equation}
If the probability of each model producing a completely consistent answer list is greater than some constant $c>0$, the expected number of samplings can be bounded by $\frac{K}{n\times c} $, 
which  reduces the number of samplings by a factor $\frac{1}{n\times c} $. 

The \textbf{expected cost savings}, representing the proportion of computation cost saved through ModelSwitch, can then be expressed as:
\begin{equation}
\text{Expected Cost Savings} = \frac{K - \mathbb{E}[N_{\text{call}}]}{K}.
\end{equation}


\section{Related Work}
Two lines of work are most relevant to ours: generation-verification paradigm and multi-agent collaboration. Generation-verification paradigm follows a two-stage pattern: generating multiple candidate answers through sampling~\cite{brown2024large,IterativePrompting}, then selecting the most suitable ones using verification mechanisms~\cite{cobbe2021gsm8k,li2022competition}. Multi-agent collaboration~\cite{talebirad2023multi,zhang-etal-2024-exploring,qian2024scaling,wan2025rema,zhang2025nature} enhances answer quality through collaborative interactions among multiple reasoning agents powered by single or multiple LLMs.

\paragraph{Generation-Verification Paradigm.} 
The generation-verification paradigm underlies most recent advances of test-time compute. 
Self-consistency~\cite{wang2022self} involves sampling various reasoning paths and selecting the most consistent answer, thereby enhancing chain-of-thought reasoning.
It is arguably the simplest generation-verification strategy, 
and has been shown to be a highly effective test-time compute strategy in LLMs. For instance, GPT-o1~\cite{o1} and Deepseek-R1~\cite{guo2025deepseek} have shown notable improvements when using self-consistency as a test-time compute strategy.  This method Recent advances enhance this paradigm through structured decomposition: Universal Self-Consistency~\cite{chen2023universal} directly generates final answers from N-best candidates via LM prompting, while Branch-Solve-Merge~\cite{saha2023branch} decomposes problems into sub-prompts and merges solutions through meta-prompting. Similarly,  Li et al.~\cite{li2024more} emphasize that increasing the number of samplings can outperform larger LLMs. Brown et al.~\cite{brown2024large} show that increasing the number of samplings consistently improves the coverage of correct answers. Our work aims to enhance generation effectiveness and efficiency by using multiple LLMs to complement each other.


Beyond basic methods like voting, verification strategies have evolved significantly:
\textit{Reward-based selection}~\cite{christiano2017deep,stiennon2020learning,kim2023prometheus,luo2024improve,zhang2024generative,cui2025process,liu2025can} significantly improves answer selection by learning to rank generated samples based on quality; \textit{Automatic verification} leverages tool calls (e.g., calculators, search engines, code executors)~\cite{li2022competition,schick2023toolformer} or mathematical proof-checking~\cite{welleck2022naturalprover} for rigorous validation; \textit{LLM-as-a-Judge}~\cite{zheng2023judging,li2024generation,lifshitz2025multi} improves verification accuracy through off-the-shelf LLMs which aligns closely with human preferences. We employ a default voting strategy that balances simplicity and effectiveness, and it has been shown to be enhanced by strong verification methods.



\paragraph{Multi-Agent Collaboration.} A recently emerging class of multi-model collaboration methods is multi-agent debate.
Multi-agent debate~\cite{du2023improving,chan2023chateval, AgentVerse, liang2023encouraging,pmlr-v235-smit24a,li2024improving,kim2024can} assigns multiple agents of the same LLM as proponents and opponents, engaging them in role-playing dialogues to refine answers through iterative debate. However, these debate methods typically rely on single homogeneous LLM. Recently, mixture of agents (MOA)~\cite{wang2024mixture,li2024smoa} refines samples by critically evaluating answers across different LLMs in iterative cycles. 
The commonality of multi-agent debate systems is that different agents (LLMs) are forced to influence each other. While showing promise, under what circumstances LLMs agree on the correct answer and when they are swayed by noise remains an open question without a well-established conclusion. 


Another class of multi-agent collaboration approach is model routing~\cite{shnitzer2023large,lu2023routing,hu2024routerbench,muqeeth2024learning,ong2024routellm}, which trains router networks to distribute questions to specialized models. Our method achieves analogous benefits through dynamic model switching based on consistency – effectively implementing a \textit{training-free router} that adaptively selects the most appropriate model per query.

Our work combines generation-verification paradigm and multi-agent collaboration by proposing a framework that leverages model diversity without relying on explicit agent interactions.

\section{Conclusion}
In this paper, we leverage the complementary strengths of multiple LLMs at the test time without requiring internal fusion or additional training to improve performance. Building on the repeated-sampling-then-voting framework, we introduce a strategy that not only enhances performance but also significantly improves computational efficiency.

Empirical observations demonstrate a strong correlation between a model's internal consistency and the accuracy of its generated final answers, laying the foundation of our approach. Extensive experiments confirm that our method ModelSwitch achieves competitive performance while substantially reducing inference costs. Theoretical analysis further validates the efficiency and performance benefits. This makes our strategy a practical and generalizable solution for various reasoning and knowledge-based tasks, paving the way for more efficient and effective applications of LLMs.



\bibliographystyle{unsrt}  
\bibliography{reference}

\newpage
\appendix
\onecolumn
\section{More Experimental Results}
\label{app:experiment}
\subsection{Experimental Setup}
\label{App:Setup}
\paragraph{Benchmarks}
The details of the seven datasets we used are as follows:
\begin{itemize}
    \item GSM8K~\cite{cobbe2021gsm8k}: A benchmark for evaluating grade school mathematical reasoning and problem-solving abilities, containing 1,319 questions.
    \item MATH~\cite{lightman2023let}: A challenging dataset consisting of 500 high school and college-level math problems.
    \item MathBench~\cite{liu2024mathbench}: A comprehensive dataset spanning a wide range of mathematical disciplines, designed to evaluate both theoretical understanding and practical problem-solving skills. We use the Arith subset, which contains 300 questions.
    \item MGSM~\cite{shi2022language}: A multi-language version of GSM8K, where we use 10 non-English languages and sample a total of 1,000 test samples.
    \item DATE~\cite{wei2022chain}: A dataset designed to test symbolic reasoning capabilities through simple string manipulation tasks, containing 396 questions.
    \item MMLU-Pro~\cite{wang2024mmlu}: A robust and challenging dataset for massive multitask understanding. We randomly select a subset of 500 questions for evaluation.
    \item AIME24: A prestigious high school mathematics competition dataset contains 30 challenging mathematical problems.
\end{itemize}
\paragraph{Hyperparameter Settings}
We set the temperature and top\_p of GPT-4o mini to 1 in all experiments, while all other LLMs were kept with their default hyperparameters.

When there are only two models, we always set external weights $\text{W}_\beta = $ uniformly to 1.
The settings for  $\text{W}_\beta$ in Sections~\ref{sec:vs.debate} and~\ref{sec:llm_number} are shown in Tables~\ref{tab:weight1} and~\ref{tab:weight2}, respectively.
\begin{table}[!h]
    \centering
    \begin{tabular}{l|cccccc}
    \hline
        Model               & GSM8K & MATH  & MathBench & MGSM  & DATE & MMLU-Pro\\
        \hline
        GPT-4o mini         & 1     & 2     & 1         & 2     & 1.5  & 1      \\
        Gemini 1.5 Flash    & 1     & 2     & 1         & 2     & 1.5  & 1.5    \\
        Claude 3 Haiku      & 1     & 1     & 1         & 1     & 1    & 1      \\
    \hline
    \end{tabular}
    \caption{Settings for external weights $\text{W}_\beta$  in Section~\ref{sec:vs.debate}.}
    \label{tab:weight1}
\end{table}
\begin{table}[!h]
    \centering
    \begin{tabular}{l|cc}
    \hline
        Model                   & MathBench & DATE \\
        \hline
        GPT-4o mini             & 2         & 1.5          \\
        Gemini 1.5 Flash        & 2         & 1.5         \\
        Claude 3 Haiku          & 2         & 1          \\
        Llama-3.1-8B-Instruct   & 1         & 1          \\
        Gemma-2-9B-It           & 1         & 1        \\
        Qwen2.5-7B-Instruct     & 1         & 1        \\
    \hline
    \end{tabular}
    \caption{Settings for external weights $\text{W}_\beta$ in Section~\ref{sec:llm_number}.}
    \label{tab:weight2}
\end{table}
\paragraph{Multi-Agent Debate Systems Setup}
For both MAD and MAD (MLD), we allocate 3 rounds of debate, with a sampling budget of 5 per round. For ChatEval, we set the collaboration round to 5, with each round involving a general public, a critic, and a scientist. For AgentVerse, we set sampling budget to 17, consisting of 3 role assigners, 5 solvers, 3 critic 0 , 3 critic 1 and 3 evaluators.
We set sampling budget of MOA to 16, consisting of 5 proposer layers with 3 proposers each and 1 aggregator layer, totaling 16 samples. For ModelSwitch, we allow three LLMs sample up to 5 times respectively.  

\paragraph{Compute Resources}
All open-source LLMs in this paper were run on a cluster running Ubuntu 22.04, equipped with 1,600GB of memory and 8 NVIDIA A100 GPUs, each with 80GB of VRAM. While this setup provides ample resources, it is not strictly necessary to run these LLMs. For smaller LLMs (not more than 10 billion parameters), full precision inference can be performed on systems with as little as two NVIDIA RTX 4090 GPUs (24GB VRAM each). For the 70B LLMs, a minimum of four NVIDIA RTX 4090 GPUs or two NVIDIA A100 GPUs (80GB VRAM each) is required.

\subsection{ModelSwitch using Reasoning LLMs on AIME24}
\label{AIME}
\begin{figure}[t!]
    \centering
    \vspace{-5pt}
    \includegraphics[width=\columnwidth]{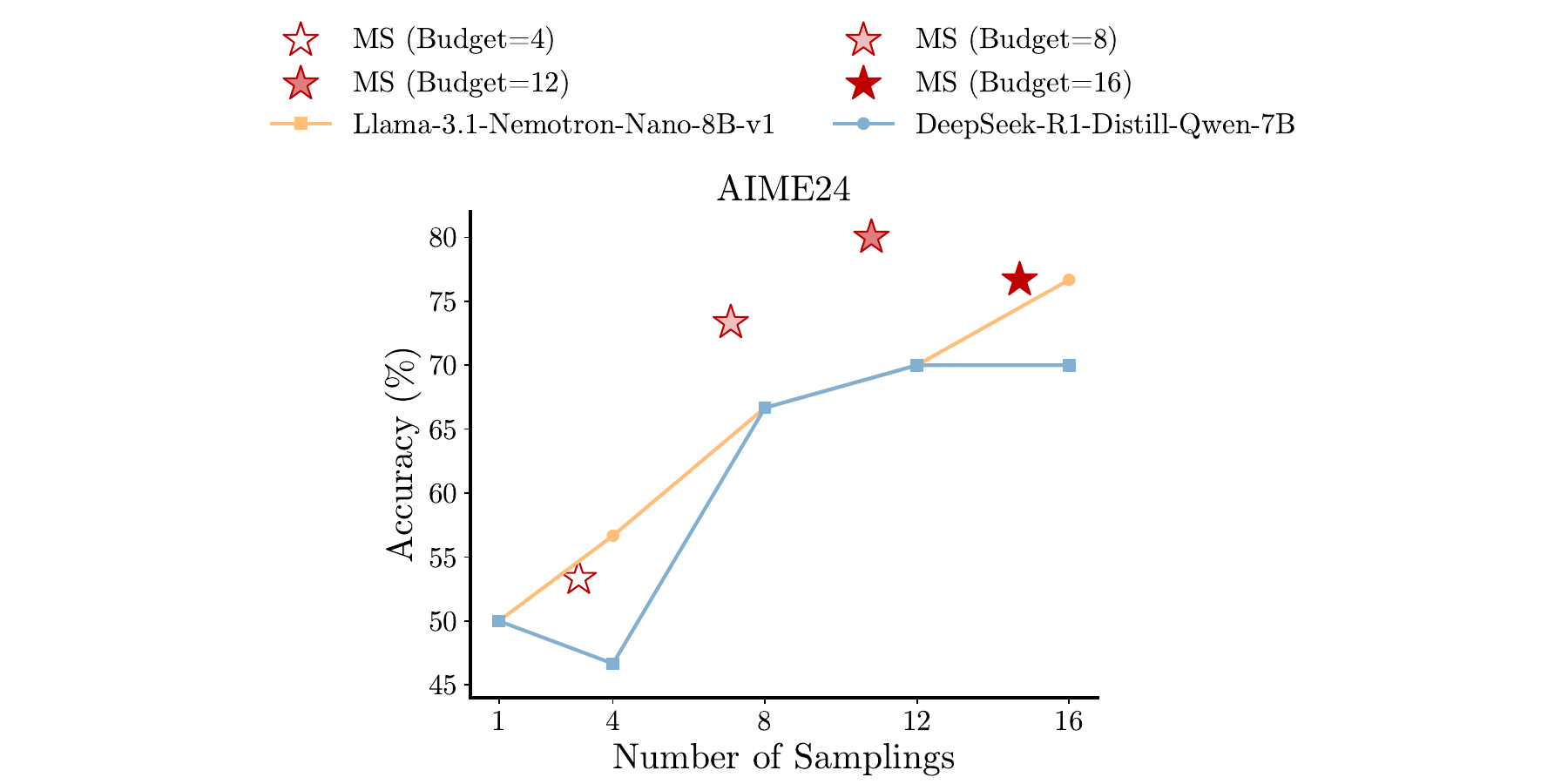}
    \caption{Performance comparison of ModelSwitch (MS) and self-consistency (SC)~\cite{wang2022self} on AIME24. MS switches between Llama-3.1-Nemotron-Nano-8B-v1 and DeepSeek-R1-Distill-Qwen-7B. The curves show SC performance for individual LLMs. ModelSwitch achieves equal or better performance with fewer samples at budget 8, 12 and 16.}
    \label{fig:reasoningLLMs}
    \vspace{-5pt}
\end{figure}

In this section, we explore whether ModelSwitch is also applicable to reasoning LLMs. We still compare ModelSwitch with self-consistency. To ensure fair comparison, we evaluate self-consistency for each LLM (Llama-3.1-Nemotron-Nano-8B-v1 and DeepSeek-R1-Distill-Qwen-7B) and ModelSwitch using both, under the same sampling budget. The results are shown in Figure~\ref{fig:reasoningLLMs}.

When the sampling budget is 8 and 12, ModelSwitch achieves significant improvements over the self-consistency of the best-performing LLM (66.7\% → 73.3\%, 70\% → 80\%). When the sampling budget is 16, ModelSwitch achieves the same accuracy (76.7\%) as the self-consistency of the best-performing LLM. The actual average sampling counts of ModelSwitch under these three budgets is 7.1, 10.8, and 14.7, respectively, saving an average of 9\% in samples (due to the difficulty of the questions and the low probability of models producing completely consistent answers, the number of samples saved is relatively limited). Overall, ModelSwitch remains an effective method for reasoning LLMs—achieving equal or better performance with fewer samples.

\subsection{Experimental Cost}
To better understand ModelSwitch's reduction in computational cost, we visually present the experimental expenses. The API call prices for GPT-4o mini and Gemini 1.5 Flash are identical: 
\$0.15 per 1M tokens for prompts and \$0.6 per 1M tokens for completions. Since the prompt (input) part is exactly the same, we only account for the cost of completions (output). Table~\ref{tab:cost} shows the actual costs for self-consistency using GPT-4o mini and Gemini 1.5 Flash, as well as for ModelSwitch using both, based on Figure~\ref{fig:main_performance} with a sampling budget of 16. Ensuring that ModelSwitch outperforms self-consistency, we also significantly reduce costs, consistent with the reduction in the number of samplings.
\begin{table}[!h]
    \centering
    \begin{tabular}{l|llllll}
    \hline
        Price (\$)               & GSM8K & MATH  & MathBench & MGSM  & DATE & MMLU-Pro\\
        \hline
        GPT-4o mini (SC)        & 2.1\textcolor[HTML]{C82423}{*}  & 0.6  & 0.3\textcolor[HTML]{C82423}{*} & 1.8     & 0.5\textcolor[HTML]{C82423}{*}  & 0.9      \\
        Gemini 1.5 Flash (SC)    & 1.4   & 0.6\textcolor[HTML]{C82423}{*}  & 0.3  & 1.4\textcolor[HTML]{C82423}{*}     & 0.3  & 0.7\textcolor[HTML]{C82423}{*}   \\
        ModelSwitch      & $1.1_{\textcolor[HTML]{32B897}{\downarrow48\%}}$    & $0.4_{\textcolor[HTML]{32B897}{\downarrow33\%}}$     & $0.2_{\textcolor[HTML]{32B897}{\downarrow33\%}}$         & $0.9_{\textcolor[HTML]{32B897}{\downarrow36\%}}$     &  $0.3_{\textcolor[HTML]{32B897}{\downarrow40\%}}$   & $0.6_{\textcolor[HTML]{32B897}{\downarrow15\%}}$      \\
    \hline
    \end{tabular}
    \caption{Cost comparison between ModelSwitch and self-consistency (SC). We use \textcolor[HTML]{C82423}{*} to indicate the best-performing LLM on each dataset, and \textcolor[HTML]{32B897}{$\downarrow$} to denote the cost reduction of ModelSwitch compared to self-consistency using the best-performing LLM. Across six datasets, we achieved significant cost reductions while ensuring performance improvements.}
    \label{tab:cost}
\end{table}
\subsection{Ablation Study}
\label{ablation study}
In this section, we investigate the improvements brought by the weighted voting algorithm in aggregating answers. As shown in Table~\ref{tab:ablation}, we conduct ablation study on the MATH, MGSM, DATE, and MMLU-Pro datasets, where both internal and external weights were used for voting. It is very clear that both internal weights and external weights are effective in voting.
\begin{table}[!h]
    \centering
    \begin{tabular}{l|llll}
    \hline
        Method               & MATH  & MGSM & DATE & MMLU-Pro\\
        \hline
        ModelSwitch         & 78.6   & 90.6  & 77.8  & 63.2      \\
        ~~~~w/o internal weights    & $78.4_{\textcolor[HTML]{32B897}{\downarrow0.2}}$ & $90.5_{\textcolor[HTML]{32B897}{\downarrow0.1}}$    & $77.0_{\textcolor[HTML]{32B897}{\downarrow0.8}}$  & $62.0_{\textcolor[HTML]{32B897}{\downarrow1.2}}$    \\
        ~~~~w/o external weights    & $77.8_{\textcolor[HTML]{32B897}{\downarrow0.8}}$  & $90.3_{\textcolor[HTML]{32B897}{\downarrow0.3}}$   & $76.4_{\textcolor[HTML]{32B897}{\downarrow1.4}}$  & $61.4_{\textcolor[HTML]{32B897}{\downarrow1.8}}$    \\
        ~~~~w/o internal and external weights    & $77.0_{\textcolor[HTML]{32B897}{\downarrow1.6}}$  & $90.2_{\textcolor[HTML]{32B897}{\downarrow0.4}}$   & $76.4_{\textcolor[HTML]{32B897}{\downarrow1.4}}$  & $61.2_{\textcolor[HTML]{32B897}{\downarrow2.0}}$    \\
    \hline
    \end{tabular}
    \caption{Ablation study of the weighted voting algorithm.}
    \label{tab:ablation}
\end{table}

\section{Difference Between Consistency and Perplexity}
The term "consistency" in this paper differs from "perplexity"~\cite{kuribayashi2021lower,mavromatis2024pack} (where perplexity is a measurement of how well a language model predicts a sample, defined as the exponential of the average negative log-likelihood of the predicted token probabilities). Instead, "consistency" refers to the agreement of generated answers produced by the model across multiple samples for queries with definitive answers. For example, the model may generate two samples with completely different reasoning processes but arrive at the same answer. In this case, two samples may have significantly different perplexities, but we consider their answers to be consistent.

\section{Theoretical Analysis for the Whole Dataset}
\label{section:Theo}
\subsection{Set Up and Notations}
\label{Set up}
To theoretically understand the promotion of the multi-LLM repeated sampling method, we first introduce a simple formulation of our problem. We start with one single query $q$ and a given model $M$. Let $M(q)$ denote the set of all possible answers of model $M$ on query $q$ and $\Delta(M(q))$ as the  distribution of $M$'s answer given query $q$. In the following context, we denote $\Delta(M(q))$ as $\Delta(M)$ for simplicity. 
Then we denote $\Phi$ as the algorithm for determining the final answer, such as majority voting. Let $\Phi(M)$ denote the distribution of model's final answers after applying $\Phi$ (we omit the query $q$). We denote $f: \Delta(M) \to \mathbb{R} $ as the loss function.  By selecting $f$ as the 0-1 loss, we can evaluate the model's performance by the probability of it generating the correct answer (we always denote $A$ as the correct answer in the following context). Then  the performance of model $M$  could be evaluated by $\mathbb{E}\big[f(\Phi(M)\big]$, where the expectation is over the query distribution.
To verify that using expectation to measure performance is reasonable, we have compared our theoretical results with the experimental results on the dataset MathBench with $K = 10,16$. The results exhibit \textbf{complete consistency} in terms of ordinal relationships and demonstrate substantial agreement at the numerical level.

Without loss of generality, we consider the case of two models. 
Let $\text{MV}$ denote the majority voting algorithm with $2K$  samples for single model, and $\text{MS}$  denote our ModelSwitch Algorithm ~\ref{alg:model_switch} samples $K$ times for model $M_1$ and model $M_2$, respectively. The performance of the these algorithms can be denoted as $\mathbb{E}\big[f(\text{MS}(M_1,M_2))\big] $, $\mathbb{E}\big[f(\text{MV}(M_1))\big] $ and $\mathbb{E}\big[f(\text{MV}(M_2))\big] $. 

In the following analysis, we show that under certain conditions, the performance of ModelSwitch can outperform single model repeated sampling then majority voting, i.e., 
\begin{equation}
 \mathbb{E}\big[f(\text{MS}(M_1,M_2)\big] < \min \Big \{ \mathbb{E}\big[f(\text{MV}(M_1))\big], \mathbb{E}\big[f(\text{MV}(M_2))\big] \Big \}. \label{eq:hx}
\end{equation}

\subsection{Two Intuitive Examples}
\label{App:examples2}
We provide two examples to aid understanding. These examples demonstrate two potential reasons why the MS method can enhance performance. We always let $f$ be the $0-1$ loss in this section.
\begin{example}
    Consider a problem set that consists of one single problem $q$ with the correct answer $A$, and two Model $M_1,M_2$; And $M_1,M_2$'s (one sample) answer's distribution are respectively $ (0.4,0.6,0)$ and $(0.4,0,0.6)$ (they share the same answer space $\{A,B,C\}$). If we let the number of samplings  $K$ be sufficiently large, by the Law of Large Numbers, we have
    \[
    f(\text{MV}(M_i)) \to 1 \;\;\;\;(i=1,2); \;\;\; f(\text{MS}(M_1,M_2)) \to 0,\;\;\;\; \text{as } K \to +\infty
    \]
    Clearly,  \cref{eq:hx} holds. 
\end{example}

We remark that, on single problem, this phenomenon does not exist in the case of binary classification.

\begin{example}
     Consider a problem set that consists of two multi-class classification problems $q_1,q_2$ with the correct answer $A$, and two Model $M_1,M_2$; And $M_1,M_2$'s (one sample) answer's distribution of the two problems are respectively: $M_1:(0.90,0.05,0.05)_{q_1}, (0.10,0.50,0.40)_{q_2}$ and $M_2: (0.10,0.50,0.40)_{q_1},(0.90,0.05,0.05)_{q_2}$, (they share the same answer space $\{A,B,C\}$).\\
    One could verify that for any $M$, ModelSwitch would outperform single model in this case, although $M_1,M_2$ has same loss on this problem set.\par
\end{example}
This example illustrates that disparities in expertise and uncertainty on single error create opportunities for model ensemble methods to leverage their potential.

\subsection{Understanding the Enhancement}
We highlight that our analysis is not confined to binary classification problems, which inherently introduces significant hardness. To the best of our knowledge, theoretical analysis of majority voting in multi-class settings are exceedingly rare, existing approach is to apply Poisson approximation~\cite{mitzenmacher2017probability} to derive an upper bound \cite{10507045}.

Informally, we first define two events:\; (1) $\Omega_{1} := \{f(\text{MV}(M_1)) < f(\text{MS}(M_1,M_2) \}$ denotes the event that ModelSwitch underperforms majority voting on this query; (2) $\Omega_{2} := \{f(\text{MV}(M_1))> f(\text{MS}(M_1,M_2) \}$ denotes the event that ModelSwitch outperforms Majority Voting on this query. Then let $\alpha = \mathbb{P}(\Omega_{1}), \beta = \mathbb{P}(\Omega_{2})$ denote the probability of these two events respectively.  The distribution of the count of each answer among the $K $ answers follows a multinomial distribution across $ K $ trials, with a probability of $\Delta(M)$. Let $x_A$ denote the number of times the right answer appears in $K$ samples.

Accordingly, we define $C^{\epsilon}(M_i)$ as the probability that  $M_i$ generates the correct answer with a frequency of at least $(1-\epsilon)$:
$
C^{\epsilon}(M_i) = \mathbb{P}(x_{A} \geq (1-\epsilon)K)).
$
Finally, we denote
$W^{\epsilon}(M_i)$ as an upper bound of the probability of $M_i$ presenting at least $(1-2\epsilon)$ frequency on a wrong answer. Using the methods of types~\cref{lem:MoTypes}, we can calculate that  $W^{\epsilon}(M_i) = (2K\epsilon)^{n+1} \cdot \{2^{K((1-2\epsilon)\log(\max_{j \neq A}\Delta(M_i)_j+\epsilon_0)}\}$.

The following theorem captures this intuitive insight: on the subset of queries where the performance discrepancy between two models is non-negligible,  if the superior model is sufficiently confident in the correct answer on average -- which is measured by $\mathbb{E}[C^{\epsilon}(M_i)|\Omega_i]$, while the inferior model exhibits adequate uncertainty among different incorrect answers -- which is measured by $(\mathbb{E}[1 - W^{\epsilon}(M_{3-i})|\Omega_i])$, then the ModelSwitch algorithm can achieve performance surpassing single model. 

More specifically, we divide the queries into two parts:(1) part I is queries that Model 2 could ‘enhance’ Model 1's performance; (2) part II is queries that Model 2 ‘mislead’ Model 1. Then, if the total ‘enhance’ effect is stronger than the total ‘mislead’ effect, ModelSwitch will outperform single model.
The condition that total ‘enhance’ effect is stronger than the total ‘mislead’ effect can be denoted as:
\begin{equation}
\begin{aligned}
\beta &\cdot\Big(\mathbb{E}[C^{\epsilon}(M_2)|
\Omega_{2}]\cdot {\mathbb{E}}[1-W^{\epsilon}(M_1)|\Omega_{2}]  -  \mathbb{E}[1 - f(\text{MV}(M_1))|\Omega_{2}] \Big)   \\
&\qquad >  \alpha \cdot \Big(\mathbb{E}[C^{\epsilon}(M_1)|\Omega_{1}]\cdot \mathbb{E}[W^{\epsilon}(M_2)|\Omega_{1}]   + {\mathbb{E}}[1-C^{\epsilon}(M_1)|\Omega_{1}] \Big), \label{eq:hx2}
\end{aligned}
\end{equation}
where $\beta,\alpha$ respectively reflects the proportion of part I and part II, while the term after $\beta$ measures the average ‘enhance’ effect and the term after $\alpha$ measures the average ‘mislead’ effect.
The detailed derivation of this condition can be found in \cref{App:proof}.
Then we have the following theorem:

\begin{theorem}
\label{thm:bigtheorem}
If there exists some $\epsilon \in [0,0.25)$ such that the condition \cref{eq:hx2} holds,  we have 
\begin{equation}
 \mathbb{E}\big[f(\text{MS}(M_1,M_2)\big] < \min \Big \{ \mathbb{E}\big[f(\text{MV}(M_1))\big], \mathbb{E}\big[f(\text{MV}(M_2))\big] \Big \} 
\end{equation}
\end{theorem}



\subsection{Validation of Theoretical Formulation}
\begin{figure}[t!]
\begin{center}  \centerline{\includegraphics[width=\columnwidth]{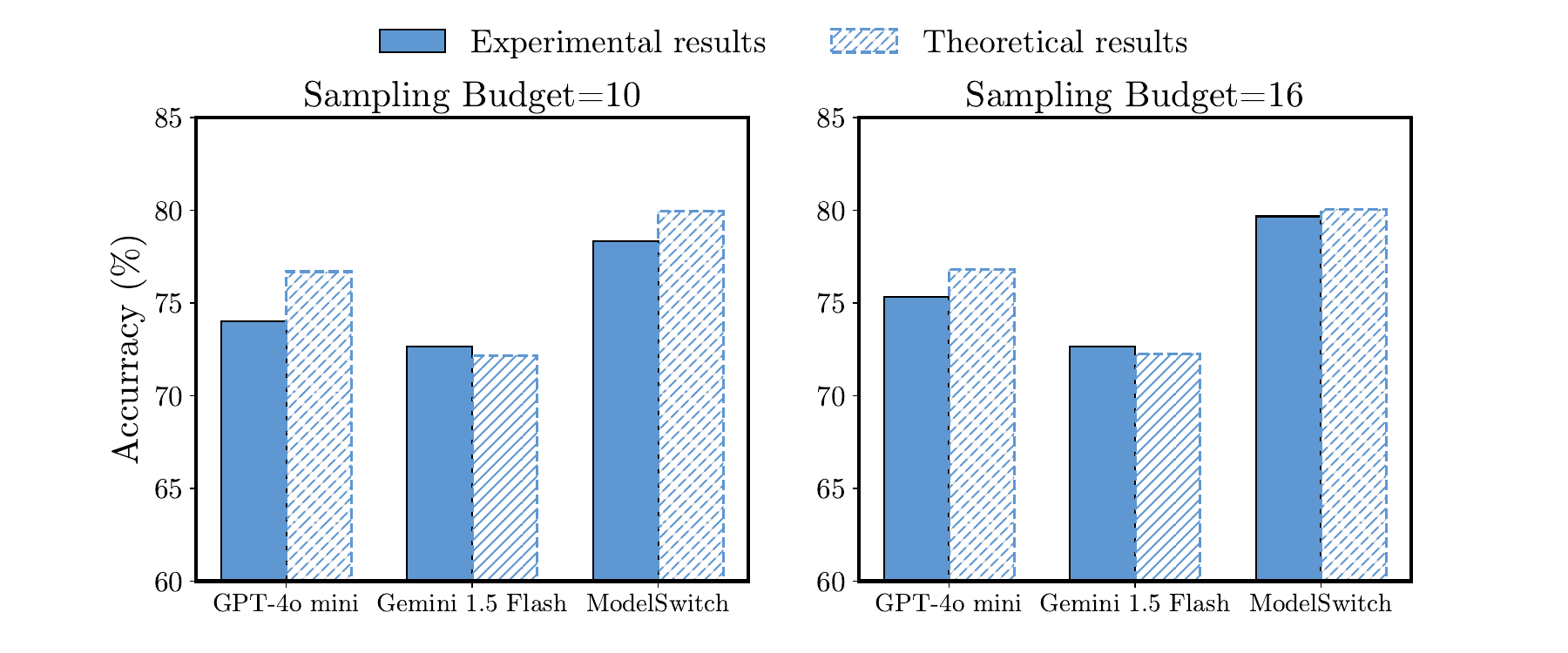}}
    \caption{Comparison of the theoretical results and the experimental results on MathBench, with sampling budgets $K= 10,16$. The theoretical results are in substantial agreement with the actual experimental results.}
    \label{App:fig-B1}
\end{center}
\end{figure}

In this subsection, 
we verify whether our formulation is reasonable. We set $f$ as $0-1$ loss and  estimate the query distribution by empirical mean. That is, for each query $q$ in the problem set, we query $M_i$  for $50$ times independently. We use  $p^{M_i}_q$ denote the answer  distribution of model $M_i$. Then use the data collected to estimate $p^{M_i}_q$ by the empirical distribution $\hat{p}^{M_i}_q$. And we view the query distribution as the uniform distribution over the whole MathBench~\cite{liu2024mathbench} problem set. In practice, we find that for each problem $q$, the total types of different answers that appear more than $1$ time is no more than $6$. So, we set $M(q) = \{A,B,C,D,E,F\}:=[A,F]$. Then, we calculate
\[
\frac{1}{N}\sum_{n=1}^{N}(\sum_{\substack{x_A = \max{x_j}\\x_A+...+x_F = M}}\frac{K!}{x_A!...x_F!}\prod_{j\in [A,F]}\hat{p}^{M_i}_{q_n}(j)^{x_j}), \;\;\;\;\; i=1,2;
\]
and
\[
\frac{1}{N}\sum_{n=1}^{N}\sum_{\substack{(\boldsymbol{x},\boldsymbol{y})\in R}}\Big(\frac{K!}{x_A!...x_F!}\Big)\cdot\Big(\frac{K!}{y_A!...y_F!}\Big)\prod_{j\in [A,F]}\big(\hat{p}^{M_1}_{q_n}(j)^{x_j}\hat{p}^{M_2}_{q_n}(j)^{y_j}\big), \;\;\;\;\; 
\]
where $R=\{(\boldsymbol{x},\boldsymbol{y})|x_A+y_A = \max_{\substack{j}}(x_j+y_j), \sum_jx_j = \sum_jy_j=K\}$. Those two expressions respectively estimate the performance of the two single model and 2-LLM repeated sampling. The results are reported in \cref{App:fig-B1}.

\subsection{Proof of Theorem C.1}
\label{App:proof}

\begin{definition}[Empirical distribution, type class]
\label{def:inj}
Let $ \boldsymbol{x} = \{x_1, \ldots, x_K\} $ with $ x_i \in \mathcal{X} = \{1, \ldots, r\} $ and  
\[N(a) = \sum_{i=1}^K \mathbbm{1}{\{x_i=a\}}, \text{P}(a) = \frac{N(a)}{K}.\]

The empirical distribution of $\boldsymbol{x}$ is denoted by $\text{P}_{\boldsymbol{x}}=(\text{P}(1), \ldots, \text{P}(r))$.  Let $\mathbb{P}_K$ denote the collection of all empirical distributions of sequences of length $ K $, i.e., 
$\mathbb{P}_K = \{\text{P}_{\boldsymbol{x}} : \boldsymbol{x} \in \mathcal{X}^K\}.$
For any $\text{P} \in \mathbb{P}_K$, the \textbf{type class} or \textbf{type} of \(\text{P}\) is denoted by
$T(\text{P}) = \{\boldsymbol{x} : \text{P}_{\boldsymbol{x}} = \text{P}\}.$
The type class of $ \boldsymbol{x} $ is 
$T_{\boldsymbol{x}} = T(\text{P}_{\boldsymbol{x}}) = \{ \tilde{\boldsymbol{x}} : \text{P}_{\tilde{\boldsymbol{x}}} = \text{P}_{\boldsymbol{x}} \}.$ Define $\mathbb{Q}$ is a probability density function, $\mathbb{Q} = \{Q(x)\}_{x\in \mathcal{X} }.$ Let  $ Q(\boldsymbol{x}) = \prod_iQ(x_i)$ and for any $S \subset \mathcal{X}^K$, 
\[
\mathbb{Q}(S) := \sum_{\boldsymbol{x} \in S}Q(\boldsymbol{x}).
\]
\end{definition}

\begin{lemma}[The Method of Types]
\label{lem:MoTypes}
\; For $\forall$ p.d.f\; $\mathbb{Q}$\; and $\forall P \in \mathbb{P}_K$,
\[
\frac{1}{(K+1)^{r-1}}2^{-K\cdot D_{\text{KL}}(P||\mathbb{Q})} \leq \mathbb{Q}(T(P)) \leq 2^{-K\cdot D_{\text{KL}}(P||\mathbb{Q})},
\]
where $D_{\text{KL}}(\cdot || \cdot)$ is the KL-divergence.
\end{lemma}
\begin{proof}
    Please refer  to~\cite{720546} for detailed proofs.
\end{proof}

\textbf{Proof of Theorem C.1}\par
\textit{Proof:} 
It is equivalent to demonstrating that the probability of ModelSwitch choosing the correct answer is higher than that of  a single model.
When $f$ is selected as $0-1$ loss, it is equivalent to show that the probability of ModelSwitch selects the correct answer (we denote it as $A$) is higher than Majority Voting on single model. Specifically, aligned with our experiment, we always set the model with lower $\mathbb{E}[f(\Phi(M))]$ to be the first model in ModelSwitch Algorithm. Thus, \cref{eq:hx} reduces to
\[
\mathbb{E}[f(\text{MS}(M_1,M_2))] <   \mathbb{E}[f(\text{MV}(M_1)].
\]
By the linearity of expectation, we have
\begin{align}
    & \mathbb{E}[f(\text{MS}(M_1,M_2)] -   \mathbb{E}[f(\text{MV}(M_1)] \nonumber \\ &= P(\Omega_{1}) \cdot \mathbb{E}[f(\text{MS}(M_1,M_2)) - f(\text{MV}(M_1)] + P(\Omega_{2}) \cdot \mathbb{E}[f(\text{MS}(M_1,M_2)) - f(\text{MV}(M_1)].\label{eq:B1}
\end{align}

Recall that $\Omega_{1} := \{f(\text{MV}(M_1)) < f(\text{MS}(M_1,M_2) \}$, $\Omega_{2} := \{f(\text{MV}(M_1))> f(\text{MS}(M_1,M_2) \}$, we know that the first term in \cref{eq:B1} is positive and the second term is negative. Next we bound them respectively.

Denote $G_K^{\epsilon}(M_i)$ as the event that among more than $(1-\epsilon)K $ samples, model $M_i$ has chosen the correct answer. Recall that
$C^{\epsilon}(M_i) = \mathbb{P}(x_{A} \geq (1-\epsilon)K) = P(G^{\epsilon}_K(M_i))$. Under the event $G_K^{\epsilon}(M_i)$, Model Switch ($2K$ samples) will always give the correct answer if the other model $M_{i'} (i' \neq i)$ generates the incorrect answers less than $(1-2\epsilon)K$ times. Moreover, by methods of types, we can derive that
\begin{align*}
\forall j\neq A,\;\;P(x_j^{M_{i'}} \geq (1-2\epsilon)K) &= \sum_{\{T(P):x_j\in{[1-2\epsilon K,K]} \}}p^{M_j}(T(P)) \\
& \leq \sum_{\{T(P):x_j\in{[1-2\epsilon K,K]} \}} 2^{-K\cdot D_{KL}(P||p^{M_{i'}})} \\
& \leq \sum_{\{T(P):x_j\in{[1-2\epsilon K,K]} \}} 2^{-K\cdot[(1-2\epsilon)\log(\frac{1}{p_j}) - \epsilon_o]} \\
& \leq (2\epsilon K)^{n+1} \cdot 2^{-K\cdot[(1-2\epsilon)\log(\frac{1}{p_j}) - \epsilon_o]},
\end{align*}
where $n = \max_{i\in\{1,2\}}\max_{\substack{q}}|\mathcal{A}_q^{M_i}|$, $\epsilon_0 = 2\epsilon(\log(2\epsilon) - \log(n-1))$. The second inequation follows from \cref{lem:MoTypes}, and the third inequation  is by  the definition of KL divergence and the last inequation is follows from the fact that eligible types are no more than $(2\epsilon K)^{n}\cdot(2\epsilon K)$.
Hence, we have 
\[
P(\{ \max_{j\neq A} \{x_j\} \geq (1-2\epsilon)K \}) \leq (2\epsilon K)^{n+1} \cdot 2^{K\cdot[(1-2\epsilon)\log(\max_{j\neq A} p_j) + \epsilon_o]}.
\]
Therefore, we have 
\begin{align}
f(\text{MS}(M_1,M_2)) - f(\text{MV}(M_1)) &\leq (1-P(G_K^{\epsilon}(M_1))) + P(G_K^{\epsilon}(M_1))\cdot P(\{\max_{j\neq A} \{x^2_j\} \geq (1-2\epsilon)K\}) \nonumber \\
&\leq (1-C_{K}^{\epsilon}(p^{M_1}))+C_K^{\epsilon}\cdot (2\epsilon K)^{n+1} \cdot 2^{-K\cdot[(1-2\epsilon)\log(\frac{1}{p_j}) - \epsilon_o]} \nonumber \\
& = (1-C_{K}^{\epsilon}(p^{M_1})) + C_K^{\epsilon}\cdot W_K^{\epsilon}(p^{M_2});\\
f(\text{MS}(M_1,M_2)) - f(\text{MV}(M_1)) & \leq
 1 - P(G_K(M_2))\cdot [1 - P(\{\max_{j\neq A} \{x^1_j\} \geq (1-2\epsilon)K\})] - f(\Phi_{MV}^{2K}(p^{M_1})) \nonumber \\
 &\leq -C_K^{\epsilon}(p^{M_2}) \cdot [1-W_K^{\epsilon}(p^{M_1})] + (1-f(\Phi_{MV}^{2K}(p^{M_1})). \label{eq:B2}
\end{align}
Note that we explicitly leverage the pattern $E[XY] = E[X\cdot E[Y|X]] \leq E[X]\cdot E[Y]$ in the last step, this inequality holds when $X,Y$ are negatively correlated or $E[Y|X]$ is decreasing with respect to $X$. This expectation decomposition inequality is natural,  since the more confidence $M_1$ puts on one single incorrect answer, the more likely $M_2$ could promote it's performance on this question. 

By choosing an $\epsilon$ that satisfies \cref{eq:hx2}, and combining \cref{eq:B1,eq:B2}, we have completed the proof of this theorem.

\end{document}